\documentclass[runningheads]{llncs}

\usepackage{eccv}

\usepackage{eccvabbrv}

\usepackage{graphicx}
\usepackage{booktabs}

\usepackage[accsupp]{axessibility}  

\usepackage{hyperref}

\usepackage{orcidlink}

\usepackage{multirow}
\usepackage{pifont}
\usepackage{pgfplots,pgfplotstable}
\pgfplotsset{set layers}
\usepgfplotslibrary{groupplots}
\usetikzlibrary{matrix,positioning}
\pgfplotsset{compat=1.18}
\usetikzlibrary{shadows}
\usepackage[dvipsnames]{xcolor}
\usepackage{adjustbox}

\begin{document}

\title{Few-Shot Synthetic Image Attribution: Identifying Unseen Generators with Limited Samples} 

\titlerunning{Few-Shot Synthetic Image Attribution}

\author{Shiyu Wu$^{*}$\inst{1,2,3} \and Shuyan Li$^{*}$\inst{4} \and Jing Li\inst{5} \and Jing Liu$^{\dagger}$\inst{1,3} \and Yequan Wang$^{\dagger}$\inst{2,6}}

\authorrunning{S. Wu et al.}

\institute{Institute of Automation, Chinese Academy of Sciences, Beijing, China \and
Beijing Academy of Artificial Intelligence, Beijing, China \and University of Chinese Academy of Sciences, Beijing, China \and Tsinghua Shenzhen International Graduate School, Tsinghua University, Shenzhen \and Harbin Institute of Technology, Shenzhen, China \and Peking University, Beijing, China \\
\email{wushiyu2022@ia.ac.cn, Lishuyan@sz.tsinghua.edu.cn, jingli.phd@hotmail.com, jliu@nlpr.ia.ac.cn, tshwangyequan@gmail.com}
}

\maketitle

\makeatletter
\renewcommand{\@makefnmark}{}
\makeatother

\footnotetext[1]{$^{*}$Equal contribution. $^{\dagger}$Corresponding Authors.}

\begin{abstract}

AI-generated image (AIGI) attribution presents a pressing challenge that goes beyond mere AIGI detection, aiming to identify the source model or technique responsible for a synthetic image. 
However, most previous source attribution methods operate in a closed-set manner, which necessitates retraining to recognize any novel category, preventing adaptation to the rapid evolution of image generation. 
In this work, we propose a new paradigm for synthetic image attribution, termed few-shot attribution. 
This paradigm targets the reliable identification of unseen generators using only limited samples, making it highly suitable for real-world applications. 
To facilitate this work, we construct OmniFake, a large-scale, well-categorized synthetic image dataset that contains $1.17$ million images from $45$ distinct generators. 
We further introduce OmniDFA (Omni Detector and Few-shot Attributor), a few-shot attribution baseline that not only assesses the authenticity of images but also determines their synthesis origins. 
Experiments demonstrate that OmniDFA exhibits excellent capability in few-shot attribution and achieves state-of-the-art generalization performance in AIGI detection. 
Our dataset and code are available at \href{https://github.com/teheperinko541/OmniDFA}{https://github.com/teheperinko541/OmniDFA}. 

\keywords{Image Attribution \and Synthetic Image Detection \and Few-Shot Learning}

\end{abstract}
    
\section{Introduction}

Generative models \cite{ddpm21} now forge photorealistic images that defy visual scrutiny, collapsing the boundary between authentic and synthetic. 
The growing threat of widespread misuse requires operational methods to not only distinguish AI-generated images (AIGI) from real ones but also trace their generative origins, which is critical for understanding model-specific vulnerabilities. 
However, existing tools for detecting and analyzing synthetic content struggle to match the pace of advancing generation methods.
This poses a significant challenge for such countermeasures in an open-set scenario, where they must handle data from generative models unseen during training, necessitating robust generalization ability and strong discrimination capability. 

\begin{figure}[t]
    \centering
    \includegraphics[width=1.0\columnwidth]{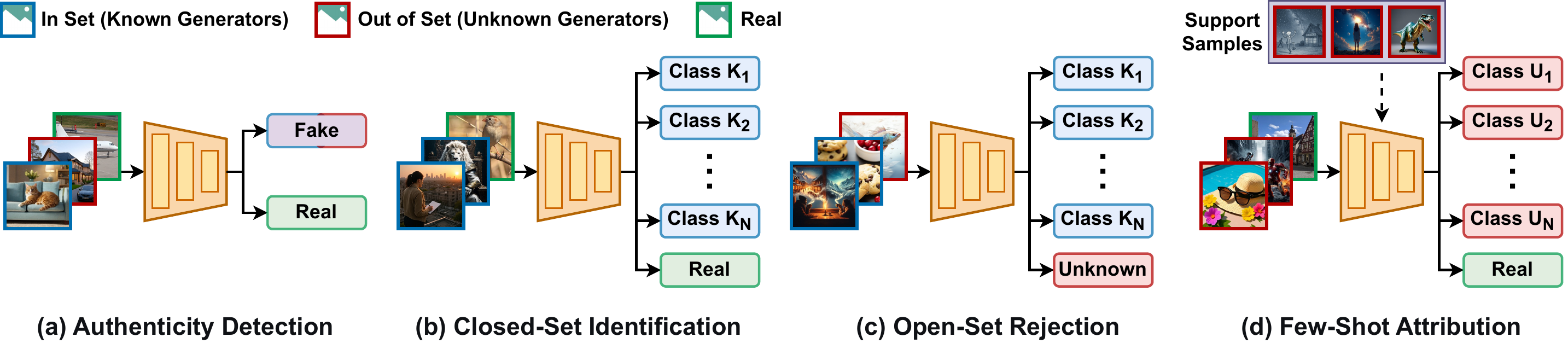}
    \caption{Comparison of task-specific pipelines for synthetic analysis. Our new attribution paradigm (d) offers dramatically improved scalability over previous works (b) and (c).}
    \label{fig:cmp_method}
\end{figure}

Synthetic image attribution represents a broader and more fine-grained task in forgery analysis, moving past the limitations of coarse-grained authenticity detection. 
It aims to identify the specific generative source among multiple candidates, consequently providing deeper insights into image provenance. 
However, existing image attribution paradigms are confined to identifying solely known generator types, rendering them ineffective against the rapidly evolving landscape of image generation. 
Specifically, closed-set methods, as illustrated in \cref{fig:cmp_method} (b), can only trace images to generators seen during the training phase, which prevents them from handling unknown sources \cite{dctcnn74}. 
In contrast, open-set methods \cite{bosc69, clipfeatures68} categorize unseen samples by collapsing all unknown sources into a single class, as shown in \cref{fig:cmp_method} (c). 
Adapting these approaches to incorporate new categories necessitates retraining the entire network, which is resource-intensive and unsustainable. 
This inherent inflexibility renders them quickly obsolete as new generative models emerge. 
The question of how to efficiently adapt to novel generators therefore represents a critical challenge for image attribution. 

In this work, we propose a new paradigm termed few-shot attribution to tackle synthetic image attribution from a novel perspective. 
This paradigm operates in a fully open-set scenario, requiring the model to identify image sources unseen during training, as depicted in \cref{fig:cmp_method} (d). 
In contrast to prior tasks, our focus lies in enhancing generalization and scalability to novel categories. 
We argue that the advantages of few-shot learning, such as rapid generalization from limited data, are also applicable to image attribution. 
Thus, we formulate this novel task, which challenges models to identify unknown generators from a minimal number of support samples. 
The model learns to rapidly extract critical artifact evidence from a limited number of samples, and once trained, it can be swiftly adapted to novel categories at test time. 
This training-free adaptation offers a substantial advantage over conventional methods: it is highly scalable, cost-effective, and well-suited for real-world deployment. 

To facilitate our work, we propose OmniFake, a well-categorized dataset specifically designed for multi-class attribution. 
We construct this dataset as our experiments require a large number of categories for both training and testing, which most previous datasets cannot provide. 
Specifically, we collect generated images from $45$ distinct generative models spanning GANs \cite{gan14}, diffusion models \cite{ddpm21}, autoregressive models \cite{showo55}, and hybrid architectures \cite{uniworld65}. 
We maintain a substantial collection of synthetic images for each class, ensuring comprehensive coverage and diversity. 
OmniFake significantly surpasses prior datasets \cite{genimage1, aigcbenchmark2} in both model diversity and up-to-date coverage, with a clear comparison presented in \cref{tb:cmp_dataset}. 
Crucially, none of the generators employ parameter-efficient fine-tuning (PEFT) techniques (e.g., DreamBooth \cite{dreambooth33} and LoRA \cite{lora32}), ensuring inherently distinct generative mechanisms rather than feature-similar variants derived from minor adaptations. 
Such diversity enriches the fundamental resources available for both AIGI detection and attribution research. 

\begin{table}[t]
    \centering
    \caption{Comparison with existing AIGI datasets. ``Attribution'' indicates that the dataset is designed for attribution task, ensuring pairwise separability among classes. }
    \small
    \scalebox{0.9}{
        \setlength{\tabcolsep}{7pt}
        \begin{tabular}{ccccc}
            \toprule
            AIGI Dataset & Year & Generators & Fake Images & Attribution \\
            \midrule
            CNNSpot \cite{cnnspot15} & CVPR 2020 & 11 & 362 K & \textcolor{green}{\ding{51}} \\
            DiffusionForensics \cite{dire29} & ICCV 2023 & 11 & 439 K & \textcolor{green}{\ding{51}} \\
            GenImage \cite{genimage1} & NeurIPS 2023 & 8 & 1.33 M & \textcolor{red}{\ding{55}} \\
            DE-FAKE \cite{defake76} & CSS 2023 & 4 & 222 K & \textcolor{green}{\ding{51}} \\
            UniversalFakeDetect \cite{univfd7} & CVPR 2023 & 19 & 400 K & \textcolor{red}{\ding{55}} \\
            Artifact \cite{artifact23} & ICIP 2023 & 25 & 2.49 M & \textcolor{red}{\ding{55}} \\
            AIGCBenchmark \cite{aigcbenchmark2} & Arxiv 2024 & 17 & 360 K & \textcolor{red}{\ding{55}} \\
            DRCT-2M \cite{drct56} & ICML 2024 & 16 & 2.00 M & \textcolor{red}{\ding{55}} \\
            ImagiNet \cite{imaginet35} & Arxiv 2025 & 8 & 100 K & \textcolor{green}{\ding{51}} \\
            WildFake \cite{wildfake5} & AAAI 2025 & 23 & 2.55 M & \textcolor{red}{\ding{55}} \\
            \midrule
            OmniFake & ECCV 2026 & \textbf{45} & 1.17 M & \textcolor{green}{\ding{51}} \\
            \bottomrule
        \end{tabular}
    }
    \label{tb:cmp_dataset}
\end{table}

Leveraging our comprehensive dataset, we propose OmniDFA (Omni Detector and Few-shot Attributor), an innovative framework that simultaneously addresses AIGI detection and open-set few-shot attribution. 
We adopt a dual-path architecture that extracts image characteristics from both low-level and high-level perspectives to effectively capture fine-grained details while maintaining global representations. 
We integrate supervised contrastive learning \cite{scl12} with center loss \cite{centerloss13} to simultaneously address both the attribution and detection tasks. 
Experimental results show that OmniDFA achieves state-of-the-art performance on our proposed OmniFake dataset, surpassing previous methods by $2.03$\% in $5$-way attribution and $5.83$\% in AIGI detection. 
Additionally, it exhibits robust zero-shot detection performance on GenImage \cite{genimage1} and Chameleon \cite{aide57}, confirming the generalizability and effectiveness of our approach. 

Our contributions are summarized as follows: 
\begin{itemize}
    \item We introduce the new task of few-shot attribution, which leverages limited samples to identify unseen generators in an open-set scenario. 
    
    \item We construct the OmniFake dataset, a large-scale, well-categorized synthetic image dataset that enables in-depth study of model-specific patterns for image attribution. 
    
    \item We propose OmniDFA, a novel framework integrating AIGI detection with few-shot attribution, establishing a strong baseline with broad applicability. 
\end{itemize}

\section{Related Work}

\subsection{Synthetic Datasets}

Early datasets for AI-generated image detection primarily relied on Generative Adversarial Networks (GANs) \cite{gan14}. 
CNNSpot \cite{cnnspot15} introduces a widely-used dataset and establishes an evaluation paradigm where detectors are trained on images from ProGAN \cite{progan16} and LSUN \cite{lsun17}, then tested across various other models.
As diffusion models \cite{ddpm21} advance in image synthesis, many datasets \cite{onlinedetection19, univfd7} have incorporated images from these models to evaluate and improve detection methods. 
GenImage \cite{genimage1} introduces the first million-scale synthetic dataset, paired with real images from ImageNet \cite{imagenet22}. However, these datasets only cover a limited number of generators, restricting the generalization capability of detectors. 
Recent studies \cite{red106-26} have begun emphasizing synthetic model diversity in dataset construction. 
WildFake \cite{wildfake5} introduces an in-the-wild collected dataset featuring diverse fake images from various generative models, covering a wide range of content and styles. 
Community Forensics \cite{communityforensics6} addresses the diversity limitation by aggregating samples from thousands of generative models. 
However, these large-scale datasets prioritize the detection task and contain numerous variants of generative models with consistent architectures. 
The high feature similarity among them can adversely affect both training and testing for the attribution task. 
Moreover, such datasets rely heavily on diffusion models, which is misaligned with the current paradigm shift toward autoregressive architectures. 

\subsection{Detection Methods}

Previous research on synthetic image detection has been dedicated to feature extraction from multiple perspectives, such as frequency \cite{frequencyaware27}, semantics \cite{c2pclip28}, and reconstruction difficulty \cite{dire29, lare30}. 
Although they have achieved promising results on previous datasets, recent studies \cite{fakeorjpeg4, safe31} suggest that data augmentation techniques such as resizing and JPEG compression can significantly degrade the performance of detectors.
To extract more robust features, OOC-CLIP \cite{oocclip36} employs the pre-trained CLIP model as its feature extractor, while FakeReasoning \cite{fakereasoning37} utilizes vision-language models, which not only effectively incorporate textual prompts but also produce human-interpretable feature descriptions. 
Recent studies \cite{code40} have introduced more sophisticated learning objectives to overcome the limitations of binary classifiers. 
DetectByReal \cite{realimagesonly38} introduces learning with real images only by analyzing pixel-level distributions. 
NTF \cite{ntf10} employs self-supervised feature mapping to enhance transfer learning, while FSD \cite{fsd9} learns a specialized metric space to distinguish unseen fake images with given samples. 
Inspired by these advances, we aim to explore the potential of metric learning not only for detection but also for out-of-distribution attribution. 

\subsection{Attribution Methods}

Image attribution aims to identify the source model of generated images. 
Early studies \cite{dctcnn74, repmix75} focus on closed-set classification over GANs, where all generators encountered during testing are assumed to be known at training time. 
They demonstrate that different generative models exhibit distinct fingerprints, which can be distinguished by a neural network. 
However, the closed-set setting has limited practicality in real-world scenarios, as new generative models are continuously being released. 
To address this limitation, the open-set rejection task has been introduced \cite{ossisa73}, where a model is required to categorize fake images from unseen generators into an additional rejection class, thereby distinguishing them from known categories. 
For instance, DNA-Det \cite{dnadet71} captures globally consistent architectural traces through patch-based contrastive learning. 
CPL \cite{pseudo70} introduces a global-local voting module to evaluate attribution performance on various GAN-generated face images. 
DE-FAKE \cite{defake76} employs the CLIP model to attribute fake images created by text-to-image diffusion models. 

Nevertheless, most existing methods are inherently limited to attributing images to generative models seen during training. 
When confronted with a newly emerging generator, they typically require retraining to adapt. 
In this work, we aim to explore rapid model adaptability to continuously emerging generative models---a practical demand that remains largely unaddressed. 
We propose an open-set few-shot paradigm to evaluate model performance in detecting forgery clues from unseen, novel generators. 
This necessitates that the model rapidly learns to identify salient features from limited samples, while also distinguishing them from features characteristic of other generative models. 

\section{Construction of OmniFake}

\begin{figure*}[t]
    \centering
    \includegraphics[width=0.96\linewidth]{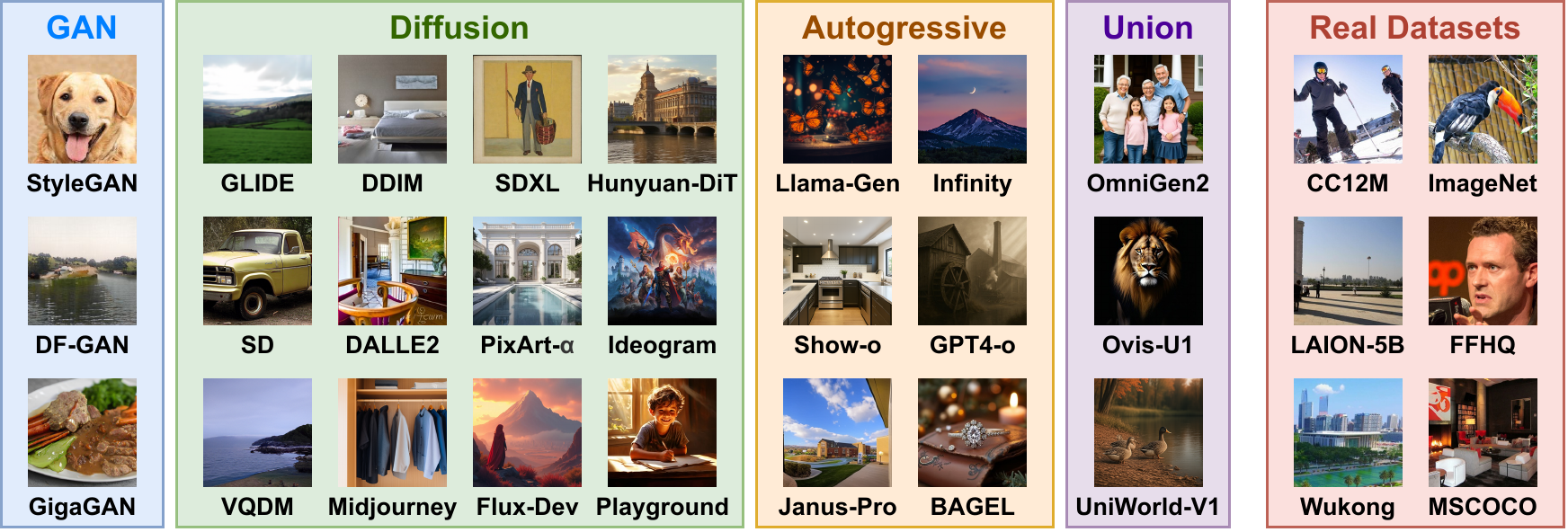}
    \caption{Samples of the OmniFake dataset. Our dataset covers a broad spectrum of generative models, with real images sourced from multiple datasets for comprehensive coverage. }
    \label{fig:datasetexamples}
\end{figure*}

As presented in \cref{tb:cmp_dataset}, most million-scale datasets do not emphasize the uniqueness of each category, collecting data from generators with similar model architectures. 
To support our investigation in multi-class attribution, we carefully construct a dataset comprising fake images sampled from a diverse set of models. 
For a visual overview, a snapshot of our dataset is shown in \cref{fig:datasetexamples}. 

\subsection{Fake Image Collection}

A key limitation of prior synthetic image datasets for attribution is the scarcity of categories, which considerably restricts research progress. 
To ensure comprehensive data diversity, we gather fake images through three distinct channels: (1) established datasets and benchmarks, (2) community-shared collections, and (3) synthetic images generated using open-source models.
We curate data from open-source datasets including GenImage \cite{genimage1}, WildFake \cite{wildfake5}, and MPBench \cite{mpbench41}. 
The final collection of this channel comprises images produced by $19$ distinct generators spanning GANs, diffusion models, and VAEs. 
To enrich our dataset with data from closed-source models, we select $8$ synthetic datasets from Hugging Face \cite{huggingface42}, such as DALLE3 \cite{dalle3_44}, Ideogram \cite{ideogram45}, and Midjourney V6 \cite{midjourney43}. 
We also select $18$ state-of-the-art open-source models from Hugging Face or the models' official repositories, and generate corresponding images based on a comprehensive collection of prompts. 
These include flow-matching models such as Hunyuan-DiT \cite{hunyuandit46} and SD3-Medium \cite{sd3_47}, as well as autoregressive models like Janus-Pro \cite{januspro48}, BAGEL \cite{bagel49}, and Show-o \cite{showo55}. 
Additionally, we incorporate several cutting-edge multimodal unified models that utilize diffusion-based decoders for high-fidelity image generation, including OmniGen2 \cite{omnigen2_50}, Ovis-U1 \cite{ovis64}, and UniWorld-V1 \cite{uniworld65}, among others. 
Details on the image generators and the synthesis pipeline are provided in the Supplementary Material. 

\subsection{Real Image Collection}

To establish a comprehensive benchmark for image forensics research, we curate a diverse collection of authentic images from $10$ publicly available datasets spanning multiple domains. 
We carefully structure our dataset into two distinct categories:
(1) Large-scale raw datasets including LAION \cite{laion400m52}, Wukong \cite{wukong53}, and CC12M \cite{cc12m54}, providing vast amounts of in-the-wild data; (2) Carefully constructed datasets such as ImageNet \cite{imagenet22} and COCO \cite{mscoco51}, offering high-quality images from specific domains. 
This approach ensures a broad and representative set of real images, facilitating robust evaluation of forensic techniques across varied contexts. 
The final curated collection covers diverse image characteristics including resolution, scene complexity, and photographic styles. 

\subsection{Analyses of OmniFake}

The OmniFake dataset comprises a total of $2.34$ M images in the training set, with a balanced distribution of $1.17$ M real images and $1.17$ M synthetic images. 
The synthetic images originate from $45$ distinct generative models that span a broad spectrum of architectures. 
This establishes a level of granularity and scale for few-shot attribution, which remains unattainable in existing benchmarks. 
To ensure sufficient intra-class diversity, we guarantee that each synthetic image category in the training set contains at least $20$ K samples. 
We also construct a separate test set comprising $90$ K synthetic images and an equal number of real images for evaluation. 
To maintain the conceptual clarity of the open-set attribution task, our dataset comprises only architecturally distinct base models, explicitly excluding fine-tuned or LoRA-adapted variants. 
We omit them to focus our evaluation on rigorous cross-structure generalization, avoiding noise from trivial parameter shifts. 
OmniFake provides a rich foundation for investigating the impact of model architectures on generalization, thereby advancing synthetic image detection and attribution. 
\section{Method}

\begin{figure*}[t]
    \centering
    \includegraphics[width=1.0\linewidth]{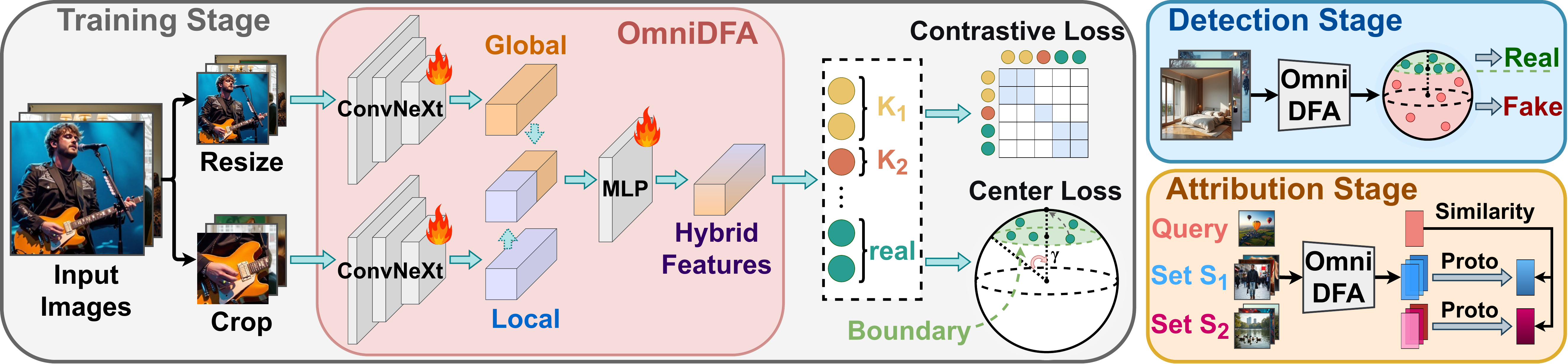}
    \caption{Overview of OmniDFA. Our dual-path architecture captures both low-level and high-level image features, balancing fine details and global representations. We use contrastive learning to enhance feature discrimination and apply sphere center loss to compactly cluster real samples. Our model is capable of tackling both the authenticity detection and image attribution tasks. }
    \label{fig:omnidfa}
\end{figure*}

In this section, we present in detail our OmniDFA, which is a novel AI-generated image (AIGI) framework that jointly addresses authenticity detection and few-shot open-set attribution. 
\cref{fig:omnidfa} illustrates the overall architecture of our proposed method.

\subsection{Few-Shot Image Attribution}

Few-shot learning provides a robust framework for our attribution task, enabling models to generalize to new categories from only a limited number of examples. 
This setting is particularly suitable for AIGI attribution, as generative models are numerous and continuously evolving, making extensive per-model data collection impractical.
Building upon this concept, we formally introduce the task of few-shot attribution, as illustrated in \cref{fig:cmp_method} (d). 

In the context of attribution, each distinct generative model is treated as a unique class, and authentic images serve as a separate reference category. 
Specifically, the attribution task is structured as an $N$-way $K$-shot problem. 
We are provided with a support set $\mathcal{S} = \cup_{i=1}^{N} \mathcal{S}_i$, which comprises samples from $N$ distinct generator classes, serving as the definitive reference library for source identification. 
Each subset $\mathcal{S}_i = \{ (\mathbf{x}_{i1}, y_{i1}), (\mathbf{x}_{i2}, y_{i2}), \ldots, (\mathbf{x}_{iK}, y_{iK}) \}$ consists of $K$ sample images produced by the $i$-th generator, where $\mathbf{x}_{ij} \in \mathbb{R}^D$ is an image with $D$ dimensions and $y_{ij} = i$ identifies the source generator. 
Given a query set of unlabeled images, the objective is to leverage the limited information in $\mathcal{S}$ to accurately attribute each query sample to its originating generator.

\subsection{Structure of OmniDFA}

Since current generative models are able to produce high-resolution images, resizing inputs to a fixed scale as in prior studies \cite{genimage1} tends to degrade fine-grained details. 
Conversely, maintaining the original resolution without resizing often results in information loss due to computational constraints or input size limitations of the model input.
To address this problem, we propose a dual sampling strategy that captures both local and global features, thereby maximizing the retention of critical visual information across varying image resolutions. 

As illustrated in \cref{fig:omnidfa}, OmniDFA processes an image $\mathbf{x}_i$ through two complementary pathways: (1) an aspect-ratio-preserving resize, where the shorter edge is scaled to the target input size followed by center cropping, to retain holistic global features; and (2) a direct high-resolution crop from the original image, which maximally preserves fine-grained textures and local details. 
The distinct global and local features from the high-level and low-level branches are channel-wise concatenated and subsequently processed through a multi-layer perceptron (MLP) to produce the final feature representation $f_\theta(\mathbf{x}_i)$, where $\theta$ denotes the set of all trainable parameters in the network.

\subsection{Unified Attribution and Detection Objective}

To facilitate both image attribution and detection, we leverage supervised contrastive learning \cite{scl12} integrated with sphere center loss, which jointly enforces angular margin maximization and feature centroid convergence. 
Specifically, for a batch of $N$ images $\{\mathbf{x}_1, \mathbf{x}_2, \ldots, \mathbf{x}_N \}$ with known categories, we first extract their features $f_\theta(\mathbf{x}_i)$ and then apply L2-normalization to obtain the corresponding vectors $\mathbf{z}_i = f_\theta(\mathbf{x}_i) / ||f_\theta(\mathbf{x}_i)||_2$. The supervised contrastive loss can be formulated as follows: 
\begin{equation}
    \mathcal{L}_{sup} = \sum_{i=1}^{N} \frac{-1}{|P(i)|} \sum_{p \in P(i)} \log \frac{e^{(\mathbf{z}_i \cdot \mathbf{z}_p / \tau)}}{\sum_{a \in P(i) \setminus \{p\}} e^{(\mathbf{z}_i \cdot \mathbf{z}_a / \tau)}},
\end{equation}
where $P(i)$ denotes the set of samples in the same class as $\mathbf{x}_i$, and $\tau$ denotes the temperature factor that scales the similarity scores.

Given that the substantial quantity of real images may lead to feature dispersion in the embedding space, we implement a sphere center loss specifically for the real category to better regularize its feature distribution, formulated as: 
\begin{equation}
    \mathcal{L}_{cen} = \frac{1}{|P_r|} \sum_{p \in P_r} (1 - \mathbf{z}_p \cdot \mathbf{c}_r),
\end{equation}
where $\mathbf{c}_r$ is a learnable L2-normalized vector representing the feature center of the real class, and $P_r$ denotes the set of all real samples in the current mini-batch. 
By unifying these constraints, we formulate the final learning objective with the scaling factor $\lambda$: 
\begin{equation}
    \mathcal{L} = \mathcal{L}_{sup} + \lambda \mathcal{L}_{cen}. 
\end{equation}

To enable direct measurement of image authenticity, we introduce a boundary threshold $\gamma$ defined as the maximum angular separation between real image features and the real center. 
We compute this bound using Tukey’s fences and update it via momentum-based adjustment, with full details provided in the Supplementary Material. 

\section{Experiment}

In this section, we first present the benchmarks involved and our experimental configurations. 
We then conduct comprehensive comparisons between OmniDFA and several state-of-the-art synthetic image detection and attribution methods. 

\subsection{Benchmarks and Evaluation Metrics}
\textbf{Datasets and benchmarks.}
We first conduct comprehensive experiments on our OmniFake dataset. 
To adapt our framework for the few-shot attribution task, we perform our experiments using $3$-fold cross-validation by randomly dividing the OmniFake dataset into three balanced parts. 
In each validation round, we perform training on two parts while using the remaining part for testing. 
The evaluation is conducted under both $5$-way $10$-shot and $15$-way $10$-shot scenarios. 
To comprehensively evaluate the detection capability of our model, we conduct extensive zero-shot experiments on the GenImage dataset \cite{genimage1} and the Chameleon dataset \cite{aide57}, demonstrating its generalization ability in both dataset-specific and real-world contexts. 

\noindent \textbf{Evaluation metrics.}
Following established practices in previous research \cite{genimage1, fsd9}, we adopt accuracy (ACC) and macro-averaged F1 score (Macro-F1) as the evaluation metrics for the attribution task. 
For the detection task, we additionally report average precision (AP), computed with a threshold step of $0.05$.  

\subsection{Experimental Settings}
We adopt the ConvNeXt-Small \cite{convnext58} pretrained on ImageNet as the feature extractor of our model, which outputs a vector of 512 dimensions. 
These features are then processed through an MLP to produce the final 128-dimensional embeddings. 
Following ComFor \cite{communityforensics6}, we employ RandAugment \cite{randaug59}, Gaussian blur, and JPEG compression during training to effectively mitigate potential biases while enhancing the robustness of our model. 

During training, we sample fake images with a per-GPU batch size of $128$ and real images with a per-GPU batch size of $16$, resulting in a total batch size of $1152$ across $8$ GPUs to meet the requirements of contrastive learning. 
We use AdamW as our optimizer with a base learning rate of $2 \times 10^{-5}$ and employ a cosine annealing learning rate scheduler for a total of $20$ epochs. 
We set the temperature parameter $\tau=0.07$ and loss coefficient $\lambda=0.01$.  
The proposed method is implemented with the PyTorch library and all the experiments are conducted on $8$ A100 with $40$ GB memory. 
Additional implementation details can be found in the Supplementary Material.

\subsection{Open-Set Few-Shot Attribution}

\begin{table*}[t]
    \centering
    \caption{Results of open-set few-shot attribution on all the three parts of OmniFake. Each task is evaluated with $10$ support samples. We report Accuracy / Macro-F1 in percentage, with the best results highlighted in boldface. }
    \small
    \scalebox{0.62}{
        \begin{tabular}{l|cc|cc|cc|cc}
            \toprule
            \multirow{3}{*}{Methods} & \multicolumn{6}{c|}{OmniFake Dataset} & \multicolumn{2}{c}{\multirow{2}{*}{Averages (\%)}} \\
            \cmidrule{2-7}
             & \multicolumn{2}{c|}{Part I} & \multicolumn{2}{c|}{Part II} & \multicolumn{2}{c|}{Part III} & & \\
             & 5-way & 15-way & 5-way & 15-way & 5-way & 15-way & 5-way & 15-way \\
             \midrule
             DNA-Det & 49.06 / 47.58 & 28.41 / 26.69 & 48.63 / 46.60 & 26.65 / 23.84 & 50.73 / 49.31 & 29.70 / 28.24 & 49.47 / 47.83 & 28.25 / 26.26 \\
             CPL & 46.54 / 44.52 & 26.82 / 24.34 & 48.42 / 46.63 & 28.09 / 26.19 & 55.73 / 53.68 & 38.31 / 34.89 & 50.23 / 48.28 & 31.07 / 28.47 \\
             SiameseNet & 45.32 / 43.83 & 28.30 / 24.89 & 50.52 / 47.65 & 30.51 / 27.18 & 51.04 / 48.95 & 36.34 / 32.17 & 48.96 / 46.81 & 31.72 / 28.08 \\
             POSE & 47.27 / 45.82 & 29.61 / 26.52 & 52.13 / 48.97 & 32.47 / 28.85 & 53.20 / 50.11 & 38.52 / 34.68 & 50.87 / 48.30 & 33.53 / 30.02 \\
             OOC-CLIP & 54.62 / 53.01 & 36.14 / 33.59 & 58.68 / 57.40 & 37.07 / 35.23 & 63.13 / 62.45 & 44.30 / 42.96 & 58.81 / 57.62 & 39.17 / 37.26 \\
             UnivAttr & 54.51 / 52.89 & 34.47 / 32.77 & 57.95 / 56.53 & 35.45 / 33.99 & 61.38 / 60.16 & 42.99 / 41.15 & 57.95 / 56.53 & 37.64 / 35.97 \\
             ComFor & 58.82 / 57.86 & 37.82 / 36.78 & 60.35 / 59.28 & 37.98 / 36.53 & 60.40 / 59.32 & 40.45 / 38.90 & 59.86 / 58.82 & 38.75 / 37.40 \\
             FSD & 67.32 / 66.07 & 42.47 / 40.88 & 75.22 / 74.53 & 53.55 / 52.40 & 77.40 / 76.96 & \textbf{60.81} / \textbf{59.93} & 73.31 / 72.52 & 52.28 / 51.07 \\
            \midrule
             OmniDFA & \textbf{69.04} / \textbf{67.93} & \textbf{44.25} / \textbf{42.09} & \textbf{78.00} / \textbf{77.19} & \textbf{56.11} / \textbf{54.39} & \textbf{78.98} / \textbf{78.24} & 60.26 / 58.73 & \textbf{75.34} / \textbf{74.45} & \textbf{53.54} / \textbf{51.74} \\
            \bottomrule
        \end{tabular}
    }
    \label{tb:fewshotcmp}
\end{table*}

The out-of-distribution classification task serves as a crucial testbed for evaluating whether a model genuinely learns essential features of unseen categories, rather than simply memorizing training patterns. 
Our study evaluates a selection of existing methods, including six attribution approaches and two synthetic image detection techniques, as summarized in \cref{tb:fewshotcmp}. 
The attribution methods are DNA-Det \cite{dnadet71}, CPL \cite{pseudo70}, SiameseNet \cite{siamese67}, POSE \cite{pose81}, OOC-CLIP \cite{oocclip36}, and UniversalAttr \cite{clipfeatures68}. 
We also include AIGI detection methods FSD \cite{fsd9} and ComFor \cite{communityforensics6}, motivated by their extensive training across a large number of categories, which suggests a strong potential for capturing diverse feature distributions. 
Although these models are not specifically designed for this task, we extract features from the last layer of their backbones and employ prototypical classification. 
Specifically, we compute the prototype centers from the support set and classify each query sample based on its distance to these centers. 
We evaluate each model on OmniFake using both $5$-way $10$-shot and $15$-way $10$-shot settings across $15$ unseen categories from each test fold. 
To ensure reliability, we conduct $10,000$ independent test episodes for each experimental configuration following the episodic testing protocol. 

As shown in \cref{tb:fewshotcmp}, our model exhibits strong few-shot learning capabilities, confirming its effectiveness in identifying subtle inter-category features. 
These characteristics make our model a powerful starting point for future research in this newly proposed task. 
FSD also demonstrates competitive performance, underscoring the effectiveness of metric learning for this task. 
To our surprise, the standard binary classifier ComFor, though untrained for multi-class tasks, inherently exhibits some classification capability. 
This stems from its extensive exposure to a wide array of models during training, which yields relatively robust feature extraction capabilities. 
Other attribution methods perform inadequately in this open-set few-shot attribution task, primarily because they focus excessively on seen categories during training and fail to generalize to unseen ones. 
Our results significantly surpass previous methods, demonstrating the effectiveness of our framework and highlighting promising directions for future research.

\subsection{Cross-Generator Authenticity Detection}

\begin{table*}[t]
    \centering
    \caption{Results of authenticity detection on OmniFake. We evaluate zero-shot detection performance across different dataset partitions, reporting both real and fake accuracy (Acc) and average precision (AP) in percentage. The best results are highlighted in boldface. }
    \small
    \scalebox{0.73}{
        \begin{tabular}{l|cccc|cccc|cccc|cc}
            \toprule
            \multirow{3}{*}{Methods} & \multicolumn{12}{c|}{OmniFake Dataset} & \multicolumn{2}{c}{\multirow{2}{*}{Averages (\%)}} \\
            \cmidrule{2-13}
             & \multicolumn{4}{c|}{Part I} & \multicolumn{4}{c|}{Part II} & \multicolumn{4}{c|}{Part III} & & \\
             & F-Acc & R-Acc & Acc & AP & F-Acc & R-Acc & Acc & AP & F-Acc & R-Acc & Acc & AP & Acc & AP \\
             \midrule
             UnivFD & 7.39 & 98.66 & 53.03 & 56.07 & 24.08 & 98.22 & 61.15 & 67.78 & 16.43 & 98.64 & 57.54 & 62.14 & 57.24 & 62.00 \\
             NPR & 74.13 & 82.18 & 78.16 & 76.98 & 75.99 & 82.21 & 79.10 & 78.38 & 79.26 & 81.93 & 80.60 & 80.35 & 79.28 & 78.57 \\
             AIDE & 77.51 & 96.28 & 86.90 & 94.01 & 83.20 & 96.32 & 89.76 & 94.62 & 78.53 & 96.20 & 87.37 & 93.68 & 88.01 & 94.10 \\
             SAFE & 76.93 & 97.76 & 87.35 & 91.46 & 73.73 & 97.61 & 85.67 & 92.01 & 65.06 & 97.67 & 81.37 & 88.94 & 84.79 & 90.80 \\
             ComFor & 75.58 & \textbf{98.96} & 87.27 & 89.78 & 80.78 & \textbf{99.06} & 89.92 & 92.50 & 85.06 & \textbf{99.07} & 92.07 & 97.10 & 89.75 & 93.13 \\
             FSD & 90.66 & 77.15 & 83.91 & - & 83.33 & 77.80 & 80.57 & - & 90.51 & 75.63 & 83.07 & - & 82.51 & - \\
            \midrule
             OmniDFA & \textbf{97.43} & 95.32 & \textbf{96.38} & \textbf{97.56} & \textbf{96.36} & 93.63 & \textbf{95.00} & \textbf{95.93} & \textbf{97.06} & 93.65 & \textbf{95.36} & \textbf{97.42} &  \textbf{95.58} & \textbf{96.97} \\
            \bottomrule
        \end{tabular}
    }
    \label{tb:detectioncmp}
\end{table*}

Our authenticity detection evaluation prioritizes the zero-shot capability of classifiers, defined as the generalization performance over previously unseen categories. 
Our experiments use the OmniFake dataset and compare against state-of-the-art methods, including UnivFD \cite{univfd7}, NPR \cite{npr8}, AIDE \cite{aide57}, SAFE \cite{safe31}, ComFor \cite{communityforensics6}, and FSD \cite{fsd9}. 
While each compared method was optimized on its respective training set, our evaluation intentionally focuses on their generalization to unseen, novel data. 
This design stems from the fundamental objective of synthetic detection, where robustness and performance beyond the training distribution are of central importance. 
Accordingly, we use their officially released weights to ensure optimal generalization. 
For thorough benchmarking against FSD based on metric learning, we retrain this model on our dataset. 
We include ComFor in our comparisons to benchmark against large-scale pretrained models. 
Although it is trained on a series of models built upon Stable Diffusion, potentially introducing dataset overlap, we consider this risk tolerable given the vast number of our testing categories. 

The results in \cref{tb:detectioncmp} show that our OmniDFA consistently outperforms prior methods in fake detection performance across all components of OmniFake, achieving an average improvement of $5.83$\%. 
This demonstrates the strong generalization capability of our model. 
We also notice that our model exhibits slightly lower accuracy in real image detection. 
This is attributed to our boundary update rule, which actively filters out anomalous data and may consequently exclude some marginal cases. 
We plan to investigate a more adaptive boundary update mechanism for better robustness in future work. 

\begin{table*}[t]
    \centering
    \caption{Zero-shot detection results on GenImage and Chameleon datasets. The best results are highlighted in boldface, while the second-optimal results are marked with underline. Our method demonstrates remarkable improvements in detection accuracy. }
    \small
    \scalebox{0.72}{
        \begin{tabular}{l|cccccccc|c|cccc}
            \toprule
             \multirow{2}{*}{Methods} & \multicolumn{9}{c|}{GenImage} & \multicolumn{4}{c}{Chameleon} \\
             \cmidrule(lr){2-10}
             & ADM & Glide & Midjourney & SD v1.4 & SD v1.5 & VQDM & Wukong & BigGAN & Average & F-Acc & R-Acc & Acc & F1 \\
            \midrule
             CNNSpot & 60.39 & 58.07 & 51.39 & 50.57 & 50.53 & 56.46 & 51.03 & 71.17 & 56.20 & 9.86 & 99.55 & 60.89 & 17.85 \\
             UnivFD & 66.87 & 62.46 & 56.13 & 63.66 & 63.49 & 85.31 & 70.93 & 95.08 & 70.49 & \textbf{85.52} & 41.56 & 60.42 & \underline{64.97} \\
             DIRE & 75.78 & 71.75 & 58.01 & 49.74 & 49.83 & 53.68 & 54.46 & 70.12 & 60.42 & 2.09 & \underline{99.73} & 57.83 & 4.08 \\
             PatchCraft & 82.17 & 83.79 & \underline{90.12} & \underline{95.38} & \underline{95.30} & 88.91 & 91.07 & \underline{95.80} & 90.32 & 1.39 & 96.52 & 55.70 & 2.62 \\
             NPR & 69.69 & 78.36 & 77.85 & 78.63 & 78.89 & 78.13 & 76.61 & 84.35 & 77.81 & 1.68 & \textbf{100.00} & 57.81 & 3.30 \\
             AIDE & \textbf{93.43} & \underline{95.09} & 77.20 & 93.00 & 92.85 & \underline{95.16} & \underline{93.55} & 83.95 & \underline{90.53} & 26.80 & 95.06 & \underline{65.77} & 40.19 \\
            \midrule
             OmniDFA & \underline{85.50} & \textbf{96.71} & \textbf{97.58} & \textbf{97.47} & \textbf{97.75} & \textbf{96.78} & \textbf{97.78} & \textbf{97.33} & \textbf{95.86} & \underline{77.46} & 88.00 & \textbf{83.48} & \textbf{80.09} \\
            \bottomrule
        \end{tabular}
    }
    \label{tb:crossdataset_cmp}
\end{table*}

Furthermore, our results indicate that training with multiple classes yields better performance than single-generator training approaches, such as UnivFD, which shows limited effectiveness when applied to out-of-distribution data. 
Another observation is that FSD demonstrates good performance in fake detection but performs poorly on real images. 
This limitation stems from its classification mechanism that relies on the nearest prototypical centroid, which becomes less effective when dealing with diverse types of fake images. 
In contrast, our model explicitly addresses the distribution of real images by incorporating a center loss term, which pulls the features of real samples closer to their class center in the embedding space, thereby enhancing the discriminative power for authentic data and significantly boosting the overall detection accuracy. 

\subsection{Cross-Dataset Evaluation}

Cross-dataset evaluation has been employed to test classifiers' generalization capability under open-set conditions \cite{communityforensics6}. 
To comprehensively assess the generalization capability of OmniDFA, we select two representative datasets: GenImage and Chameleon, which serve as benchmarks for standard experimental evaluation and real-world application scenarios, respectively. 
To rigorously ensure experimental reliability, we train an additional classifier by explicitly excluding all generator categories in both benchmarks and their related model families. 
For instance, we completely remove SD 1.5, SDXL, and SD3-Medium generated samples from our training data to ensure a strictly zero-shot testing condition. 
We incorporate three additional baseline methods: CNNSpot \cite{cnnspot15}, DIRE \cite{dire29}, and PatchCraft \cite{aigcbenchmark2}.  
All comparative models are developed following the experimental protocol of AIDE \cite{aide57}. 
Since the Chameleon dataset is imbalanced, we additionally report the corresponding F1 score. 

As shown in \cref{tb:crossdataset_cmp}, OmniDFA achieves remarkably high performance in these challenging zero-shot scenarios. 
Our model achieves state-of-the-art classification accuracy on the GenImage benchmark, further validating the effectiveness of our multi-generator training strategy. 
The results reveal that the single-generator-trained generalization approach, which remains prevalent in synthetic image detection, is increasingly inadequate for addressing evolving challenges. 
More importantly, on the Chameleon benchmark, our model surpasses the second-best method by a significant margin of $17.71$\% in accuracy. 
It maintains balanced performance across both authentic and fake images, demonstrating superior generalization and robust practical applicability, whereas most existing models exhibit a strong systematic bias. 
Therefore, our work represents a significant step towards more generalizable detection models. 

\subsection{Ablation Studies}

\begin{figure}[t]
  \centering
\begin{subfigure}{0.49\linewidth}
\begin{adjustbox}{max width=\linewidth}
\begin{tikzpicture}

\pgfplotsset{
    myaxis/.style={
        width=4.7cm,
        height=4cm,
        grid=none,
        xtick=data,
        font=\small,
        tick label style={font=\small},
    }
}

\begin{groupplot}[
    group style={
        group size=2 by 1,
        x descriptions at=edge bottom,
    },
    myaxis
]

\nextgroupplot[
        title={5-way},
        xlabel={Number of samples},
        ylabel={Accuracy (\%)},
        xtick={1,3,5,10,25,50,200}, 
        xmode=log,
        log ticks with fixed point,
        ytick={50,60,70,80},
        ymin=45,
        ymax=85,
        ylabel shift={-2mm},
    ]

\addplot[mark=*, mark options={scale=1},
          color={rgb,255:red,89;green,169;blue,79}, thick]
    coordinates {(1,59.57) (3,69.88) (5,72.86) (10,75.34) (25,76.91) (50,77.96) (200, 78.53)};
\label{plots:omnidfa}
\addplot[mark=triangle*, mark options={scale=1.25}, color={rgb,255:red,242;green,192;blue,164}, thick]
    coordinates {(1,53.90) (3,66.13) (5,71.24) (10,73.31) (25,74.84) (50,76.43) (200, 77.29)};
\label{plots:fsd}

\nextgroupplot[
        title={15-way},
        xlabel={Number of samples},
        xtick={1,3,5,10,25,50,200}, 
        xmode=log,
        log basis x=2,
        log ticks with fixed point,
        ytick={20,30,40,50,60},
        ymin=25,
        ymax=65,
        ylabel shift={-2mm},
    ]

\addplot[mark=*, mark options={scale=1},
          color={rgb,255:red,89;green,169;blue,79}, thick]
    coordinates {(1,37.63) (3,46.82) (5,50.26) (10,53.54) (25,55.41) (50,56.54) (200,56.98)};
\addplot[mark=triangle*, mark options={scale=1.25}, color={rgb,255:red,242;green,192;blue,164}, thick]
    coordinates {(1,31.08) (3,42.57) (5,46.91) (10,52.28) (25,53.86) (50,54.92) (200, 55.73)};
\end{groupplot}

\tikzset{
    legend box/.style={
        draw=gray!60,
        fill=white,
        rounded corners=3pt,
        drop shadow={
            shadow xshift=2pt,
            shadow yshift=-2pt,
            opacity=0.3,
            fill=gray!50
        },
        thick
    }
}

\matrix[
    matrix of nodes,
    anchor=north west,
    legend box,
    inner ysep=0.6em,
    font=\footnotesize,
    column sep=1ex,
    row sep=0ex,
    nodes={anchor=west}
  ] at ([yshift=-13mm, xshift=12.5mm] group c1r1.south west)
  {
    \ref*{plots:omnidfa} OmniDFA \quad
    \ref*{plots:fsd} FSD \\
  };

\end{tikzpicture}
\end{adjustbox}
\caption{Influence of the number of shots}
\label{fig:ablationshot}
\end{subfigure}
\hfill
\begin{subfigure}{0.49\linewidth}
\begin{adjustbox}{max width=\linewidth}
\begin{tikzpicture}

\pgfplotsset{
    myaxis/.style={
        width=4.7cm,
        height=4cm,
        grid=none,
        xtick=data,
        font=\small,
        tick label style={font=\small},
    }
}

\begin{groupplot}[
    group style={
        group size=2 by 1,
        x descriptions at=edge bottom,
    },
    myaxis
]

\nextgroupplot[title={JPEG Compression},
        xlabel={Quality},
        ylabel={Accuracy (\%)},
        xtick={95,85,75,65,55}, 
        ytick={50,60,70,80,90,100},
        ymin=45,
        ymax=102.5,
        x dir=reverse,
        ylabel shift={-2mm},
        extra x ticks={75},
        extra x tick style={
            grid=major,
            grid style={ultra thick, gray!70, dashed},
            tick label style={},
        }]
\addplot[mark=*, mark options={scale=1}, color={rgb,255:red,89;green,169;blue,79}, thick]
    coordinates {(100,96.38) (95,96.34) (85,97.12) (75,96.05) (65,94.80) (55,93.60)};
\label{plots:omni}
\addplot[mark=triangle*, mark options={scale=1.25}, color={rgb,255:red,114;green,147;blue,203}, thick]
    coordinates {(100,87.27) (95,84.37) (85,80.30) (75,77.17) (65,73.35) (55,71.26)};
\label{plots:comf}
\addplot[mark=triangle*, mark options={scale=1.25,rotate=180}, color={rgb,255:red,237;green,125;blue,49}, thick]
    coordinates {(100,86.90) (95,54.26) (85,56.84) (75,59.22) (65,58.74) (55,58.81)};
\label{plots:aide}
\addplot[mark=square*, mark options={scale=1}, color={rgb,255:red,168;green,120;blue,110}, thick]
    coordinates {(100,87.35) (95,50.59) (85,50.18) (75,50.17) (65,50.31) (55,50.45)};
\label{plots:safe}
\addplot[mark=diamond*, mark options={scale=1.25}, color={rgb,255:red,82;green,84;blue,163}, thick]
    coordinates {(100,78.16) (95,67.52) (85,50.94) (75,48.37) (65,48.99) (55,48.89)};
\label{plots:npr}

\nextgroupplot[title={Gaussian Blur}, 
        xlabel={\large{$\sigma$}},
        xtick={0,1,2,3,4}, 
        ytick={50,60,70,80,90,100},
        ymin=45,
        ymax=102.5,
        ylabel shift={-2mm},
        extra x ticks={2},
        extra x tick style={
            grid=major,
            grid style={ultra thick, gray!70, dashed},
            tick label style={},
        }]
\addplot[mark=*, mark options={scale=1}, color={rgb,255:red,89;green,169;blue,79}, thick]
    coordinates {(0,96.38) (1,95.59) (2,93.04) (3,81.19) (4,63.45)};
\addplot[mark=triangle*, mark options={scale=1.25}, color={rgb,255:red,114;green,147;blue,203}, thick]
    coordinates {(0,87.27) (1,87.56) (2,88.33) (3,86.66) (4,82.30)};
\addplot[mark=triangle*, mark options={scale=1.25,rotate=180}, color={rgb,255:red,237;green,125;blue,49}, thick]
    coordinates {(0,86.90) (1,83.63) (2,61.94) (3,53.63) (4,52.05)};
\addplot[mark=square*, mark options={scale=1}, color={rgb,255:red,168;green,120;blue,110}, thick]
    coordinates {(0,87.35) (1,70.53) (2,52.79) (3,51.61) (4,51.14)};
\addplot[mark=diamond*, mark options={scale=1.25}, color={rgb,255:red,82;green,84;blue,163}, thick]
    coordinates {(0,78.16) (1,76.89) (2,70.08) (3,60.81) (4,57.24)};

\end{groupplot}

\tikzset{
    legend box/.style={
        draw=gray!60,
        fill=white,
        rounded corners=3pt,
        drop shadow={ 
            shadow xshift=2pt,
            shadow yshift=-2pt,
            opacity=0.3,
            fill=gray!50
        },
        thick
    }
}

\matrix[
    matrix of nodes,
    anchor=north west,
    legend box,
    inner sep=0.3em,
    font=\footnotesize,
    column sep=1ex,
    row sep=0ex,
    nodes={anchor=west}
  ] at ([yshift=-13mm, xshift=3mm] group c1r1.south west)
  {
    \ref*{plots:omni} OmniDFA
    \ref*{plots:comf} ComFor
    \ref*{plots:aide} AIDE  \\
    \ref*{plots:safe} SAFE
    \ref*{plots:npr}  NPR    \\
  };

\end{tikzpicture}
\end{adjustbox}
    \caption{Robustness to compression and blurring}
    \label{fig:transformations}
\end{subfigure}
\caption{Ablation studies on the number of support shots and robustness to real-world degradations. The vertical dotted line in (b) marks the augmentation threshold. }
\end{figure}

In this subsection, we conduct ablation studies on experimental settings and our model. 
We investigate the impact of sample size and data perturbations, and evaluate the effectiveness of each component. 

\noindent \textbf{Impact of the number of shots.}
To investigate the impact of the number of support samples, we evaluate two models, OmniDFA and FSD, varying the number of support samples per class from $1$ to $200$. 
For a comprehensive analysis, we report the mean accuracy across the entire OmniFake dataset.
The summarized results are presented in \cref{fig:ablationshot}, where the horizontal axis is on a logarithmic scale. 
The results show that accuracy increases rapidly from $1$-shot to $10$-shot, after which further improvements become more gradual. 
This demonstrates the validity of our few-shot attribution task, in which only a small number of unseen examples are needed to achieve strong open-set identification performance. 
Our results suggest that $10$-shot represents a favorable trade-off between high performance and computational cost. 

\noindent \textbf{Robustness to real-world degradation.}
Image corruptions such as JPEG compression are commonly encountered during real-world image transmission and distribution.
We apply JPEG compression and Gaussian blurring to examine how image quality affects model performance. 
\cref{fig:transformations} reports the accuracy on OmniFake part I under various perturbation strengths, and the vertical dotted line indicates the augmentation boundary adopted during our training. 
The results demonstrate that our model maintains robust performance within the augmentation range and exhibits strong resilience to JPEG compression even beyond the augmentation range used in training. 
In contrast, other methods suffer significant performance degradation from the outset. 
Even when trained with JPEG augmentation, ComFor remains susceptible to this influence. 
When it comes to Gaussian blurring, our model performs excellently within the trained augmentation range, yet its performance declines noticeably beyond it. 
ComFor demonstrates excellent performance in handling blur because it is trained with extensive and targeted blur augmentation. 
This observation underscores the importance of incorporating such perturbations during training, thus providing strong empirical support for future model development. 

\begin{figure*}[t]
    \centering

\pgfplotstableread{
clu A      B      C      D      E
1   96.95  94.41  91.89  91.66  92.86
2   94.20  91.49  92.53  82.80  97.22
3   95.58  92.94  92.21  87.23  95.04
4   96.97  95.92  96.86  0     96.54
}\dataDetection

\pgfplotstableread{
clu A      B      C      D      E 
5   72.86  68.32  71.38  70.55  56.71
6   50.24  44.89  50.01  47.92  38.26
}\dataAttribution

\begin{tikzpicture}
\begin{groupplot}[
    group style={
        group size=2 by 1,
        horizontal sep=15mm,
    },
    height=3.5cm,
    ybar=1pt,
    /pgf/bar width=7pt,
    xtick=data,
    xtick pos=bottom,
    ytick pos=left,
    scaled y ticks=false,
    yticklabel style={/pgf/number format/fixed},
    x tick label style={anchor=north},
    legend style={
        anchor=north,
        legend columns=3,
        column sep=1ex,
        draw=none,
    },
    area legend,
    legend image code/.code={
    \draw[#1] (0cm,-0.1cm) rectangle (0.4cm,0.1cm);
    },
    every axis legend/.append style={font=\scriptsize},
]

\nextgroupplot[
    ylabel={Value (\%)},
    ylabel shift={-2mm},
    width=0.64\textwidth,
    ymin=77.5, ymax=102.5,
    ytick={80,85,90,95,100},
    xticklabels={F-Acc,R-Acc,Acc,AP},
    enlarge x limits=0.18,
]
\addplot[fill=NavyBlue!70]  table [x=clu, y=A] {\dataDetection};
\addplot[fill=Peach!70]   table [x=clu, y=B] {\dataDetection}; 
\addplot[fill=JungleGreen!70] table [x=clu, y=C] {\dataDetection}; 
\addplot[fill=Goldenrod!70]table [x=clu, y=D] {\dataDetection}; 
\addplot[fill=Thistle!70]  table [x=clu, y=E] {\dataDetection}; 

\node[above=0pt] at (axis cs:2.5,97) {Detection};

\nextgroupplot[
    ylabel={Accuracy (\%)},
    ylabel shift={-1mm},
    xticklabels={5-way,15-way},
    width=0.36\textwidth,
    ymin=35, ymax=85, 
    ytick={40,50,60,70,80},
    enlarge x limits=0.52,
    x=1.53cm,
    legend to name=grouplegend,
]
\addplot[fill=NavyBlue!70]  table [x=clu, y=A] {\dataAttribution};
\addplot[fill=Peach!70]   table [x=clu, y=B] {\dataAttribution};
\addplot[fill=JungleGreen!70] table [x=clu, y=C] {\dataAttribution};
\addplot[fill=Goldenrod!70] table [x=clu, y=D] {\dataAttribution};
\addplot[fill=Thistle!70] table [x=clu, y=E] {\dataAttribution};

\node[above=0pt] at (axis cs:5.5,74) {Attribution};

\addlegendentry{OmniDFA}
\addlegendentry{w/o local branch}
\addlegendentry{w/o global branch}
\addlegendentry{w/o center loss}
\addlegendentry{w/o contrastive learning}

\end{groupplot}

\tikzset{
    legend box/.style={
        draw=gray!60,
        fill=white,
        rounded corners=3pt,
        drop shadow={
            shadow xshift=2pt,
            shadow yshift=-2pt,
            opacity=0.3,
            fill=gray!50
        },
        thick
    }
}

\path (group c1r1.south) -- (group c2r1.south)
      node[legend box, below=1mm of current bounding box.south] {\ref*{grouplegend}};

\end{tikzpicture}

\caption{Ablation study on model component. Our dual-branch architecture significantly outperforms its single-branch counterpart. Moreover, contrastive learning enables our model to capture features that are more discriminative for unseen samples. }
    \label{fig:compcmp}
\end{figure*}

\noindent \textbf{Effectiveness of our modules.}
To investigate the impact of each component, we perform a series of ablation studies. 
We analyze the effects of removing the local and global branches, as well as of removing the center loss and boundary threshold. 
Additionally, we train a binary classifier without employing contrastive learning. 
We train and evaluate all models under the same setup, yielding the average detection and attribution performance across all three parts of the OmniFake dataset, as shown in \cref{fig:compcmp}. 
The results demonstrate the clear advantage of our dual-path model over the single-branch variants, confirming the superiority of our multi-level feature fusion. 
Our OmniDFA outperforms a plain binary classifier by capturing finer inter-class distinctions, which consequently enhances fake image detection accuracy. 
The results demonstrate that our model establishes a strong baseline for the novel paradigm we have introduced, thereby facilitating future research. 

\section{Conclusion}

In this paper, we establish a new paradigm named few-shot synthetic image attribution, which aims to identify unknown image generators with only a limited number of samples. 
We present OmniFake, a comprehensively curated dataset that is both sufficiently large and meticulously categorized, enabling detailed investigation into model-specific artifacts. 
Building upon this foundation, we propose OmniDFA (Omni Detector and Few-shot Attributor), a unified framework based on few-shot learning that jointly performs authenticity detection and few-shot source attribution. 
Experiments show that OmniDFA not only achieves state-of-the-art generalization in synthetic image detection, but also exhibits strong open-set attribution capability with limited reference samples. 
The results underscore the real-world applicability of our approach, suggesting a promising path for future research.

\section*{Acknowledgements}

This research is supported by Artificial Intelligence-National Science and Technology Major Project (2023ZD0121200), and the National Natural Science Foundation of China (62437001, 62436001, 62531026), and the Natural Science Foundation of Jiangsu Province under Grant BK20243051, and the Strategic Priority Research Program of Chinese Academy of Sciences under Grant XDB1350103.

%
%
\bibliographystyle{splncs04}
\bibliography{main}

\appendix

\section*{\Large Appendix}

\section{Necessity of Well-Categorized Dataset}

Most currently available datasets fail to account for the architectural uniqueness of generative models. 
For instance, the widely-used GenImage dataset \cite{genimage1} includes three models: Stable Diffusion 1.4, Stable Diffusion 1.5, and Wukong, all of which share identical backbone structures. 
This structural homogeneity results in remarkably similar feature distributions among their generated images.
Such design leads to significant limitations: even a simple CNN classifier trained on images from any one model demonstrates exceptionally strong generalization performance when evaluated on the other two models \cite{genimage1}. 
This architectural redundancy severely compromises the objectivity and comprehensiveness of model evaluation. 
Additionally, another study \cite{wildfake5} demonstrates that this generalization capability extends to models produced via fine-tuning \cite{dreambooth33} and LoRA adaptation \cite{lora32}. 
Therefore, merely making minor modifications to the weights or model architecture is insufficient to produce fundamentally different representations. 

To comprehensively evaluate the generalization ability of deepfake detectors, we argue that models across all categories must possess distinct forgery features. 
A straightforward approach is to utilize models with different architectures, as empirical evidence shows that CNNs can effectively distinguish between these categories. 
To address these requirements and support our experimental objectives, we introduce OmniFake, a large-scale comprehensive dataset comprising images synthesized by a diverse collection of distinct generative models. 
This dataset not only enables systematic investigation of detection generalization across architectures, but also facilitates out-of-distribution classification tasks. 
We believe our dataset will advance research in the fields of both synthetic image detection and attribution.

\section{Collection of Fake Images}

\begin{table*}[t]
    \centering
    \caption{Collections of generators for fake images in our OmniFake dataset. }
    \renewcommand{\arraystretch}{1.25}
    \scalebox{0.57}{
    \begin{tabular}{|c|c|c|c|c|}
    \hline
    \textbf{Generator} & \textbf{Training} & \textbf{Test} & \textbf{Source} & \textbf{Link} \\
    \hline
    
    ADM  & 30k & 2k & \multirow{5}{*}{GenImage} & \multirow{5}{*}{\url{https://github.com/GenImage-Dataset/GenImage}} \\
    GLIDE  & 30k & 2k &  &  \\
    Midjourney V5  & 30k & 2k &  &  \\
    Stable Diffusion V1.5  & 30k & 2k &  &  \\
    VQDM  & 30k & 2k &  &  \\
    \hline
    DALLE2  & 30k & 2k & \multirow{12}{*}{WildFake} & \multirow{12}{*}{\url{https://github.com/hy-zpg/AIGC-Image-Detection-Dataset}} \\
    StyleGAN3  & 30k & 2k &  &  \\
    DF-GAN  & 30k & 2k &  &  \\
    GALIP  & 30k & 2k &  &  \\
    GigaGAN & 25k & 2k &  &  \\
    DDIM & 30k & 2k &  &  \\
    DDPM & 30k & 2k &  &  \\
    Imagen & 30k & 2k &  &  \\
    Midjourney V4 & 30k & 2k &  &  \\
    SDXL & 30k & 2k &  &  \\
    VQVAE & 30k & 2k &  &  \\
    Muse & 30k & 2k &  &  \\
    \hline
    IF & 30k & 2k & \multirow{2}{*}{Fake Image Dataset} & \multirow{2}{*}{\url{https://huggingface.co/datasets/InfImagine/FakeImageDataset}} \\
    Cogview2 & 20k & 2k &  &  \\
    \hline
    FLUX-dev & 30k & 2k & \multirow{8}{*}{Hugging Face} & \url{https://huggingface.co/datasets/lehduong/flux_generated} \\
    GPT4-o & 30k & 2k &  & \url{https://huggingface.co/datasets/yufan/GPT4O_Image_T2I} \\
    DALLE3 & 30k & 2k &  & \url{https://huggingface.co/datasets/OpenDatasets/dalle-3-dataset} \\
    Phoenix & 30k & 2k &  & \url{https://huggingface.co/datasets/bigdata-pw/leonardo/} \\
    PixArt-Alpha & 27k & 2k &  & \url{https://huggingface.co/datasets/PixArt-alpha/PixArt-Eval30K} \\
    Playground V2.5 & 30k & 2k &  & \url{https://huggingface.co/datasets/bigdata-pw/playground} \\
    Ideogram & 30k & 2k &  & \url{https://huggingface.co/datasets/terminusresearch/ideogram-75k} \\
    Midjourney V6 & 30k & 2k &  & \url{https://huggingface.co/datasets/terminusresearch/midjourney-v6-520k-raw} \\
    \hline
    DiT-XL/2 & 23k & 2k & \multirow{18}{*}{Self-synthesized} & \url{https://github.com/facebookresearch/DiT} \\
    Janus-Pro & 24k & 2k &  & \url{https://github.com/deepseek-ai/Janus} \\
    BAGEL & 23k & 2k &  & \url{https://github.com/ByteDance-Seed/Bagel} \\
    OmniGen & 23k & 2k &  & \url{https://huggingface.co/Shitao/OmniGen-v1} \\
    SD3-Medium & 23k & 2k &  & \url{https://huggingface.co/stabilityai/stable-diffusion-3-medium} \\
    Hunyuan-DiT & 23k & 2k &  & \url{https://huggingface.co/Tencent-Hunyuan/HunyuanDiT-v1.2-Diffusers} \\
    Show-o & 23k & 2k &  & \url{https://github.com/showlab/Show-o} \\
    LUMINA-Image 2.0 & 23k & 2k &  & \url{https://huggingface.co/Alpha-VLLM/Lumina-Image-2.0} \\
    SANA V1.5 & 20k & 2k &  & \url{https://huggingface.co/Efficient-Large-Model/SANA1.5_4.8B_1024px_diffusers} \\
    CogView4 & 20k & 2k &  & \url{https://huggingface.co/zai-org/CogView4-6B} \\
    OmniGen2 & 20k & 2k &  & \url{https://github.com/VectorSpaceLab/OmniGen2} \\
    HiDream-I1-Dev & 20k & 2k &  & \url{https://github.com/HiDream-ai/HiDream-I1} \\
    Infinity & 20k & 2k &  & \url{https://github.com/FoundationVision/Infinity} \\
    Llama-Gen & 20k & 2k &  & \url{https://github.com/FoundationVision/LlamaGen} \\
    UniWorld-V1 & 20k & 2k &  & \url{https://github.com/PKU-YuanGroup/UniWorld-V1} \\
    BLIP3-o & 20k & 2k &  & \url{https://github.com/JiuhaiChen/BLIP3o} \\
    BRIA3.2 & 20k & 2k &  & \url{https://huggingface.co/briaai/BRIA-3.2} \\
    Ovis-U1 & 20k & 2k &  & \url{https://github.com/AIDC-AI/Ovis-U1} \\
    \hline
    \end{tabular}
    }
    \label{tb:collectionofmodels}
\end{table*}

\subsection{Fake Image Composition}
Our dataset comprises synthetic images collected through three primary sources to ensure diversity and comprehensiveness. 
First, we incorporate images from established open-source datasets and benchmarks, including GenImage \cite{genimage1}, WildFake \cite{wildfake5}, and MPBench \cite{mpbench41}, covering a wide range of generative models such as GANs, diffusion models, and VAEs. This subset consists of images generated by $19$ distinct generators, providing a solid foundation of varied synthetic content. 
We filter out fine-tuned or modified generative models to maintain clear distinctions among the categories. 
Second, we augment our dataset with community-shared collections from platforms like Hugging Face \cite{huggingface42}, integrating synthetic images from $6$ closed-source models and $2$ popular community models. 
With the rapid advancement of generative models, many previously state-of-the-art architectures have become outdated in terms of both design and performance. 
To ensure our study reflects the latest progress in the field, we source $18$ cutting-edge open-source models from Hugging Face or the official repositories. 
We generate synthetic images using these models to ensure coverage of diverse architectures such as flow-matching models, autoregressive models and unified MLLMs. 
We summarize the key details of these models in \Cref{tb:collectionofmodels}.

Our OmniFake contains $1.17$ M synthetic images generated by 45 distinct generative models, spanning a broad spectrum of architectures. 
Our synthetic categories include multiple models from the same family, but we rigorously ensure that they are not derived from the same backbone. 
To ensure both diversity and detectability, each synthetic category includes at least $20$ K training samples generated from a wide distribution of text prompts. 
For comprehensive evaluation, we provide a separate test set of $90$ K synthetic images, with $2$ K samples per category. 
Our dataset features highly structurally diverse generator categories, along with the latest and top-tier generators. 
OmniFake contains images with a minimum resolution of $200
\times200$ pixels, while most are $512\times512$ or higher.
This comprehensive collection provides a robust foundation for investigating the impact of model architectures on generalization. 

\subsection{Custom Data Synthesis}
To ensure the diversity and richness of prompts in our self-synthetic categories, we sample text descriptions from multiple text-image datasets or benchmarks, covering different granularities of textual inputs. 
Specifically, we adopt BLIP-3o-60k \cite{blip3o79} for fine-grained, highly detailed captions, CC12M \cite{cc12m54} for coarse-grained and concise prompts, and Laion-COCO \cite{laion400m52} for in-the-wild descriptions. 
To maintain representativeness and diversity in prompt selection, we balance the sampling ratio among these sources at $2:5:3$, effectively integrating their respective characteristics. 
Based on these curated prompt lists, we leverage $17$ distinct generative models to synthesize images, maximizing variety in the generated outputs. 
To ensure optimal image generation quality, we adhere to the official recommended resolutions. 
When multiple resolutions are provided, we randomly select among them to maintain diversity in the output image dimensions. 
For DiT-XL/2, which only accepts class labels as conditional input, we randomly sample from the categories for image generation. 
Although this approach may introduce certain domain biases, we argue that the pronounced artifacts inherent to the synthetic generators could potentially overshadow such biases. 
Ultimately, our multi-source, multi-granularity prompt sampling strategy ensures broad coverage and high quality in the synthetic dataset. 

\section{Collection of Real Images}
Prior image forensics benchmarks \cite{cnnspot15, genimage1} often rely on a single source dataset to represent authentic images.
However, this practice can introduce significant real biases by limiting the diversity and representativeness of the authentic class.
To mitigate this limitation and establish a more robust and generalizable benchmark, we follow WildFake \cite{wildfake5} and curate our authentic image collection by strategically sampling from a diverse set of $10$ publicly available datasets, spanning multiple domains and collection paradigms. 
An overview of the source datasets is provided in \Cref{tb:collectionofreal}. 

\begin{table}[t]
    \centering
    \caption{Collection of datasets for real images in our OmniFake dataset. }
    \setlength{\tabcolsep}{12pt}
    \begin{tabular}{|c|c|c|}
        \hline
        \textbf{Real Image Dataset} & \textbf{Training} & \textbf{Test} \\
        \hline
        Laion-5B & 251k & 20k \\
        \hline
        Wukong & 242k & 20k \\
        \hline
        ImageNet-1k & 174k & 15k \\
        \hline
        CC12M & 160k & 15k \\
        \hline
        MSCOCO & 113k & 10k \\
        \hline
        FFHQ & 68k & 2k \\
        \hline
        CelebA-HQ & 28k & 2k \\
        \hline
        LSUN-church & 80k & 2k \\
        \hline
        IMD2020 & 33k & 2k \\
        \hline
        FODB & 21k & 2k \\
        \hline
    \end{tabular}
    \label{tb:collectionofreal}
\end{table}

Our authentic image dataset is constructed from two complementary categories of publicly available sources to ensure both broad coverage and high quality. 
Large-scale raw datasets such as LAION \cite{laion400m52}, Wukong \cite{wukong53}, and CC12M \cite{cc12m54} provide essential in-the-wild diversity by capturing the unfiltered heterogeneity of real-world internet imagery at scale, which is crucial for evaluating forensic models under realistic conditions. 
These are balanced with carefully constructed datasets including ImageNet \cite{imagenet22}, MSCOCO \cite{mscoco51} and several other representative collections, which offer domain-specific focus through their manually curated collections of high-quality images representing well-composed photographic contexts. 
To conserve resources and avoid dataset conflicts, we strategically source our LAION and Wukong samples from the training set of WildFake, while drawing ImageNet samples from the GenImage training collection. 

We allocate varying proportions based on dataset size and our requirements, obtaining $1.17$ million authentic images for our training set. We also select $90$ K real images matching the size of the fake images to serve as our test set.
By integrating these different components into dataset construction, we aim to achieve a balanced representation of both uncontrolled web content and professionally captured images.

\section{Calculation of Decision-Boundary}
Since our model is built upon the new task of few-shot attribution, we aim to apply it to detection in a simple yet effective way. 
During training, we incorporate center loss to cluster the features of real samples around a specific point in the feature space. 
A statistical upper bound is then derived from this cluster to maximize generalization, with any sample residing beyond this boundary being classified as synthetic. 
We compute this bound using Tukey’s fences and update it via momentum-based adjustment. 

For a batch of $N$ real images $\{\mathbf{x}_1, \mathbf{x}_2, \ldots, \mathbf{x}_N\}$, we first extract and normalize their features: 
\begin{equation}
\mathbf{z}_i = \frac{f_\theta(\mathbf{x}_i)}{||f_\theta(\mathbf{x}_i)||_2},
\end{equation}
where $\theta$ denotes the set of all trainable parameters in our network. 
Then we compute the angle between each real sample and the real class center $\mathbf{c}_r$: 
\begin{equation}
    \theta_i = \arccos(\mathbf{z}_i \cdot \mathbf{c}_r). 
\end{equation}
We employ angular distance as the metric instead of cosine similarity because the feature distribution of real samples around the center typically exhibits an approximately symmetric angular distribution. 
In contrast, the nonlinear compression inherent to cosine similarity causes the distribution to become excessively dense near the center, thereby reducing sensitivity for discrimination. The samples are then arranged in ascending order of angular distance: 
\begin{equation}
    0 < \theta_{j_1} \leq \theta_{j_2} \leq \ldots \leq \theta_{j_N} < \pi, 
\end{equation}
where $j_k$ denotes the original index of the sample at the $k$-th position in the sorted order. 
Then we calculate quartiles for Tukey's fences using the sorted data. 
Specifically, we determine the first quartile ($Q_1$) and the third quartile ($Q_3$), which correspond to the $25$th and $75$th percentiles of the distribution, respectively: 
\begin{equation}
    Q_1 = \theta_{j_\frac{N}{4}},
\end{equation}
\begin{equation}
    Q_3 = \theta_{j_\frac{3N}{4}}.
\end{equation}
Here we assume that $N$ is divisible by $4$. 
The interquartile range (IQR) is then derived as 
\begin{equation}
    IQR = Q_3 - Q_1. 
\end{equation}
This robust measure of statistical dispersion allows us to establish a preliminary boundary for the current batch. 
According to Tukey's method, any data point lying above $Q_3 + 1.5 \times IQR$ is considered a potential outlier. 
Thus, we compute the batch-specific threshold as 
\begin{equation}
    \gamma_{b} = Q_3 + 1.5 \times IQR, 
\end{equation}
which frames our outlier detection criterion to identify samples that are anomalously far from the real center. 

During training, we apply a momentum-based update to the global boundary threshold $\gamma$ using the batch-specific estimate $\gamma_{b}$. 
This is implemented as an exponential moving average:
\begin{equation}
\gamma^- \leftarrow \beta\gamma^- + (1 - \beta) \gamma_{b}, 
\end{equation}
where $\beta$ is a momentum coefficient. 
This mechanism stabilizes the boundary against the inherent volatility of mini-batch statistics. 
The global threshold $\gamma$ can gradually evolve over time to reflect long-term changes in the feature distribution as the model's parameters $\theta$ are updated during training. 
This approach effectively decouples the short-term noise within a batch from the long-term trend of the feature distribution, resulting in a robust and smoothly evolving decision boundary that is essential for reliable outlier detection.

\section{Experiments Settings}

\begin{table*}[t]
    \centering
    \caption{Data split of the OmniFake dataset. }
    \begin{tabular}{|c|>{\centering\arraybackslash}p{10cm}|}
        \hline
        \textbf{Split Part} & \textbf{Generators Included} \\
        \hline
         Part I & Hunyuan-DiT, Imagen, SANA V1.5, DF-GAN, Janus-Pro, DDPM, Midjourney V5, OmniGen2, FLUX-dev, BRIA3.2, Ovis-U1, Cogview2, VQVAE, Phoenix, DiT-XL/2\\
         \hline
         Part II & OmniGen, LUMINA-Image 2.0, Show-o, SD3-Medium, Midjourney V6, GALIP, LlamaGen, Ideogram, Infinity, Muse, StyleGAN3, ADM, IF, GigaGAN, VQDM\\
         \hline
         Part III & SDXL, Playground V2.5, UniWorld-V1, BLIP3-o, Midjourney V4, CogView4, PixArt-Alpha, DALLE2, DDIM, GLIDE, GPT4-o, HiDream-I1-Dev, BAGEL, DALLE3, Stable Diffusion V1.5\\
         \hline
    \end{tabular}
    \label{tb:splitdataset}
\end{table*}

\begin{table}[t]
  \centering
  \caption{Data augmentation configurations during training. }
  \setlength{\tabcolsep}{12pt}
  \begin{tabular}{cc}
    \toprule
    Augmentation & Settings \\
    \midrule
    random JPEG & p=0.5, quality=(75, 95) \\
    resize & scale=(0.5, 2.0) \\
    horizontal flip & p=0.5 \\
    RandAugment & p=0.5, magnitude=9, layers=2 \\
    gaussian blur & p=0.5, sigma=(0.1, 2.0) \\
    normalize & [0,1] \\
    \bottomrule
  \end{tabular}
  \label{tb:dataaugment}
\end{table}

\begin{table}[t]
  \centering
  \caption{Training configuration for our OmniDFA. }
  \setlength{\tabcolsep}{12pt}
  \begin{tabular}{cc}
    \toprule
    Hyperparameter & Value \\
    \midrule
    model & ConvNeXt-Small \\
    MLP hidden dims & 512 \\
    MLP out dims & 128 \\
    input resolution & 3$\times$224$\times$224 \\
    batch size (fake, per GPU) & 128 \\
    batch size (real, per GPU) & 16 \\
    total epochs & 20 \\
    warmup epochs & 2  \\
    optimizer & AdamW \\
    scheduler & CosineAnnealing \\
    learning rate & $2 \times 10^{-5}$ \\
    min learning rate & 0 \\
    weight decay & $1 \times 10^{-2}$ \\
    precision & bfloat16 \\
    world size & 8 \\
    $\tau$ & 0.07 \\
    $\lambda$ & 0.01 \\
    $\beta$ & 0.99 \\
    \bottomrule
  \end{tabular}
  \label{tb:settings}
\end{table}

\subsection{Split of OmniFake}
To ensure a fair evaluation in our zero-shot task, we employ a $3$-fold cross-validation strategy on the OmniFake dataset. 
The dataset is randomly divided into three mutually exclusive and balanced parts, as shown in \Cref{tb:splitdataset}. 
In each validation round, two folds are used for training, while the remaining fold serves as the held-out test set. 
This approach not only maximizes the utilization of limited data but also guarantees that model performance is evaluated across diverse and representative subsets. 

\begin{table}[t]
  \centering
  \caption{Test configuration for few-shot attribution. }
  \setlength{\tabcolsep}{12pt}
  \begin{tabular}{cc}
    \toprule
    Hyperparameter & Value \\
    \midrule
    N-way & 5, 15 \\
    K-shot & 10 \\
    N-query & 5  \\
    episodes num & 10,000 \\
    \bottomrule
  \end{tabular}
  \label{tb:testsettings}
\end{table}

\subsection{Implementation Details}

To guarantee the reproducibility of our experiments, we present comprehensive training details, summarized in \Cref{tb:dataaugment} and \Cref{tb:settings}. 
In \Cref{tb:dataaugment}, $p$ denotes the probability of applying the corresponding transformation. 
Notably, in RandAugment \cite{randaug59}, we deliberately exclude shear and translate transformations to ensure that our local feature extractor does not capture artifacts from padding regions beyond image boundaries. 
The data augmentation is only employed during the training phase. 
The configuration for evaluating on the few-shot attribution task is shown in \Cref{tb:testsettings}. 
In our cross-dataset validation experiments, we adopt the following hyperparameters: $40$ training epochs and a $\lambda$ value of $5 \times 10^{-3}$, which are carefully selected to enhance the robustness of our results.
Our implementation is carried out using PyTorch library, with all experiments executed on a cluster of 8 NVIDIA A100 GPUs. 

\section{Impact of Image Degradation} 

\begin{figure*}[t]
    \centering
    \begin{adjustbox}{max width=0.96\linewidth}
\begin{tikzpicture}

\pgfplotsset{
    myaxis/.style={
        width=6cm,
        height=5.5cm,
        grid=none,
        xtick=data,
        font=\small,
        tick label style={font=\small},
        legend columns=-1,
        legend style={draw=none, font=\small},
    }
}

\begin{groupplot}[
    group style={
        group size=3 by 1,
        horizontal sep=1.2cm,
        x descriptions at=edge bottom,
    },
    myaxis
]

\nextgroupplot[title={JPEG},
        xlabel={Quality},
        ylabel={F-Acc},
        xtick={95,85,75,65,55}, 
        x dir=reverse,
        ylabel shift={-2mm},
        extra x ticks={75},
        extra x tick style={
            grid=major,
            grid style={ultra thick, gray!70, dashed},
            tick label style={},
        }]
\addplot[mark=*, mark options={scale=1}, color={rgb,255:red,89;green,169;blue,79}, thick]
    coordinates {(100,97.43) (95,96.24) (85,95.48) (75,93.05) (65,90.46) (55,87.7)};
\label{aplots:omni}
\addplot[mark=triangle*, mark options={scale=1.25}, color={rgb,255:red,114;green,147;blue,203}, thick]
    coordinates {(100,75.58) (95,69.74) (85,61.38) (75,55.04) (65,47.26) (55,43.03)};
\label{aplots:comf}
\addplot[mark=triangle*, mark options={scale=1.25,rotate=180}, color={rgb,255:red,237;green,125;blue,49}, thick]
    coordinates {(100,77.51) (95,10.98) (85,14.94) (75,19.51) (65,18.32) (55,18.34)};
\label{aplots:aide}
\addplot[mark=square*, mark options={scale=1}, color={rgb,255:red,168;green,120;blue,110}, thick]
    coordinates {(100,76.93) (95,2.28) (85,1.02) (75,0.91) (65,1.08) (55,1.29)};
\label{aplots:safe}
\addplot[mark=diamond*, mark options={scale=1.25}, color={rgb,255:red,82;green,84;blue,163}, thick]
    coordinates {(100,74.13) (95,52.85) (85,14.63) (75,6.58) (65,5.11) (55,4.78)};
\label{aplots:npr}

\nextgroupplot[title={Gaussian Blur}, 
        xlabel={$\sigma$},
        ylabel={F-Acc},
        xtick={0,1,2,3,4}, 
        ytick={70,80,90,100},
        ymin=67.5,
        ymax=102.5,
        ylabel shift={-2mm},
        extra x ticks={2},
        extra x tick style={
            grid=major,
            grid style={ultra thick, gray!70, dashed},
            tick label style={},
        }]
\addplot[mark=*, mark options={scale=1}, color={rgb,255:red,89;green,169;blue,79}, thick]
    coordinates {(0,97.43) (1,97.83) (2,98.02) (3,99.38) (4,99.79)};
\addplot[mark=triangle*, mark options={scale=1.25}, color={rgb,255:red,114;green,147;blue,203}, thick]
    coordinates {(0,75.58) (1,76.03) (2,78.34) (3,79.45) (4,79.84)};
\addplot[mark=triangle*, mark options={scale=1.25,rotate=180}, color={rgb,255:red,237;green,125;blue,49}, thick]
    coordinates {(0,77.51) (1,93.35) (2,97.10) (3,99.05) (4,98.39)};
\addplot[mark=square*, mark options={scale=1}, color={rgb,255:red,168;green,120;blue,110}, thick]
    coordinates {(0,76.93) (1,89.58) (2,99.39) (3,99.88) (4,99.93)};
\addplot[mark=diamond*, mark options={scale=1.25}, color={rgb,255:red,82;green,84;blue,163}, thick]
    coordinates {(0,74.13) (1,78.13) (2,88.58) (3,93.34) (4,94.14)};

\nextgroupplot[title={Gaussian Blur}, 
        xlabel={$\sigma$},
        ylabel={R-Acc},
        xtick={0,1,2,3,4}, 
        ytick={0,50,100},
        ylabel shift={-2mm},
        extra x ticks={2},
        extra x tick style={
            grid=major,
            grid style={ultra thick, gray!70, dashed},
            tick label style={},
        }]
\addplot[mark=*, mark options={scale=1}, color={rgb,255:red,89;green,169;blue,79}, thick]
    coordinates {(0,95.32) (1,93.35) (2,88.05) (3,62.99) (4,27.11)};
\addplot[mark=triangle*, mark options={scale=1.25}, color={rgb,255:red,114;green,147;blue,203}, thick]
    coordinates {(0,98.96) (1,99.10) (2,98.31) (3,93.86) (4,84.75)};
\addplot[mark=triangle*, mark options={scale=1.25,rotate=180}, color={rgb,255:red,237;green,125;blue,49}, thick]
    coordinates {(0,96.28) (1,73.90) (2,26.77) (3,8.20) (4,5.71)};
\addplot[mark=square*, mark options={scale=1}, color={rgb,255:red,168;green,120;blue,110}, thick]
    coordinates {(0,97.76) (1,51.48) (2,6.18) (3,3.34) (4,2.34)};
\addplot[mark=diamond*, mark options={scale=1.25}, color={rgb,255:red,82;green,84;blue,163}, thick]
    coordinates {(0,82.18) (1,75.65) (2,51.58) (3,28.27) (4,20.33)};

\end{groupplot}

\tikzset{
    legend box/.style={
        draw=gray!60,
        fill=white,
        rounded corners=3pt,
        drop shadow={
            shadow xshift=2pt,
            shadow yshift=-2pt,
            opacity=0.3,
            fill=gray!50
        },
        thick
    }
}

\matrix[
    matrix of nodes,
    anchor=east,
    legend box,
    inner sep=0.3em,
    font=\small,
    column sep=1ex,
    row sep=1.5ex,
    nodes={anchor=south}
  ] at ([yshift=-15mm, xshift=-10mm] group c3r1.south)
  {
    \ref*{aplots:omni} OmniDFA \quad
    \ref*{aplots:comf} ComFor \quad
    \ref*{aplots:aide} AIDE \quad
    \ref*{aplots:safe} SAFE \quad
    \ref*{aplots:npr} NPR \\
  };
\end{tikzpicture}
\end{adjustbox}
    \caption{Robustness to compression and blurring on fake or real images only. The vertical lines indicate the perturbation bounds used during our training. }
    \label{fig:transformations2}
\end{figure*}

To further investigate the impact of compression and blurring on synthetic image detection, we apply these perturbations to both fake and real images and evaluate the detection performance using F-Acc (accuracy on fake images) and R-Acc (accuracy on real images). 
Since most real images are already in JPEG format, no additional compression is applied or studied for them. 
The results shown in \Cref{fig:transformations2} reveal that most existing detection models fail quickly with only slight JPEG compression, often misclassifying compressed synthetic images as real. 
This suggests that previous datasets may lack sufficient compression diversity. 
Interestingly, under Gaussian blurring, the F-Acc score increases with stronger blur intensity. 
This indicates a non-negligible distinction between synthetic and real images. 
Therefore, incorporating diverse perturbations during training is highly necessary. 
These findings provide valuable insights and highlight a clear direction for future research.
\end{document}